\documentclass{article}

\usepackage{arxiv}

\usepackage[utf8]{inputenc}             
\usepackage[T1]{fontenc}                 
\usepackage[colorlinks]{hyperref}       
\usepackage{url}                        
\usepackage{booktabs}                   
\usepackage{amsfonts}                   
\usepackage{nicefrac}                   
\usepackage{microtype}                  
\usepackage{lipsum}
\usepackage{graphicx}

\usepackage{amsmath}
\usepackage{subcaption}
\usepackage{adjustbox}
\usepackage{multirow}
\usepackage{authblk}

\usepackage{threeparttable}
\usepackage{changepage}

\graphicspath{ {./figs/} }

\title{WACA-UNet: Weakness-Aware Channel Attention for Static IR Drop Prediction in Integrated Circuit Design}

\author[1]{Youngmin Seo}
\author[1]{Yunhyeong Kwon}
\author[3]{Younghun Park}
\author[2]{HwiRyong Kim}
\author[3]{Seungho Eum}
\author[1]{Jinha Kim}
\author[2]{Taigon Song}
\author[3]{Juho Kim}
\author[1,3,\thanks{Corresponding author.}]{Unsang Park} 

\affil[1]{Department of Artificial Intelligence, Sogang University, Seoul, South Korea\newline
\{ymin98, kyh97910, jhkmo510, unsangpark\}@sogang.ac.kr
}
\affil[2]{School of Electronic and Electrical Engineering, Kyungpook National University, Daegu, South Korea\newline
\{tgb0221, tsong\}@knu.ac.kr
}  
\affil[3]{Department of Computer Science, Sogang University, Seoul, South Korea\newline
\{zeropark, seunghoeum, jhkim\}@sogang.ac.kr
}

\date{}

\begin{document}

\maketitle

\begin{abstract}
Accurate spatial prediction of power integrity issues, such as IR drop, is critical for reliable VLSI design. However, traditional simulation-based solvers are computationally expensive and difficult to scale. We address this challenge by reformulating IR drop estimation as a pixel-wise regression task on heterogeneous multi-channel physical maps derived from circuit layouts. Prior learning-based methods treat all input layers (e.g., metal, via, and current maps) equally, ignoring their varying importance to prediction accuracy. 
To tackle this, we propose a novel Weakness-Aware Channel Attention (WACA) mechanism, which recursively enhances weak feature channels while suppressing over-dominant ones through a two-stage gating strategy. Integrated into a ConvNeXtV2-based attention U-Net, our approach enables adaptive and balanced feature representation. 
On the public ICCAD-2023 benchmark, our method outperforms the ICCAD-2023 contest winner by reducing mean absolute error by 61.1\% and improving F1-score by 71.0\%. These results demonstrate that channel-wise heterogeneity is a key inductive bias in physical layout analysis for VLSI.
\end{abstract}


\section{Introduction}
    As feature sizes approach the single-digit-nanometre regime, the power-delivery network (PDN) has emerged as a primary bottleneck of performance and yield in modern integrated circuit (IC) designs~\cite{kadagala2023iccad}. The PDN consists of multiple metal layers interconnected by vias, forming a complex hierarchical grid structure that distributes power from package pins to individual transistors. Each metal layer exhibits different sheet resistance characteristics. Lower layers typically have higher resistance due to narrower geometries, while upper layers provide lower-resistance global distribution paths~\cite{kozhaya2002multigrid}.
    
    \begin{figure*}[ht]
        \hfill 
        \begin{subfigure}{\textwidth}
            \centering
            \includegraphics[width=0.7\textwidth]{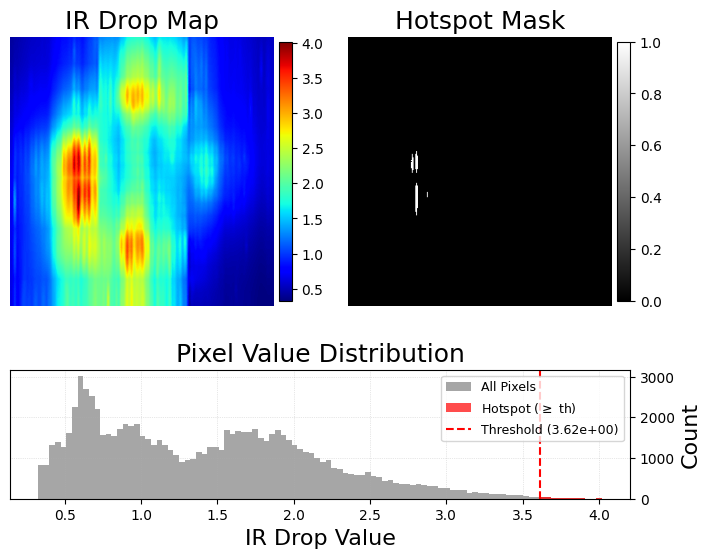}
        \end{subfigure}
        \caption{
        Visualization of the IR drop prediction task.
        Top left: Golden IR drop map for a validation sample.
        Top right: Hotspot mask highlighting regions where IR drop values exceed 90\% of the map's maximum value.
        Bottom: Histogram of pixel values in the IR drop map, with the hotspot region (IR drop $\geq$ 0.9 $\times$ maximum) highlighted in red and the threshold indicated by a dashed line.}
        \label{fig:intro-irdrop}
    \end{figure*}
    
    Among the various power-integrity phenomena, IR drop represents the voltage loss along resistive PDN paths under current flow and directly degrades timing slack, potentially triggering functional failures. IR drop severity is particularly pronounced in high-performance processors and graphics processing units, where peak current densities can exceed several amperes per square millimeter~\cite{chhabria2022openpdn}. The heterogeneous nature of multi-layer PDNs creates complex current distribution patterns, where current flows preferentially through lower-resistance paths while creating potential hotspots in high-resistance regions. As illustrated in Fig.~\ref{fig:intro-irdrop}, IR drop maps exhibit highly non-uniform spatial distributions, with critical hotspots concentrated in localized areas that require careful consideration during design optimization.
    
    Conventional static IR drop analysis solves large sparse linear systems derived from modified nodal analysis (MNA), modeled as $GV = J$, where $G$ represents the conductance matrix, $V$ the voltage vector, and $J$ the current source vector~\cite{zhao2002hierarchical,zhong2005fast}. For modern billion-node PDNs with intricate multi-layer topologies, such simulations require hours to days on high-end servers~\cite{chhabria2022openpdn}. This poses a significant challenge that can delay tape-out schedules, forcing designers to iterate with reduced accuracy, downsampled layouts, or region-specific analysis~\cite{ajayi2019openroad}, all of which risk missing late-stage IR drop violations.
    
    To mitigate these computational bottlenecks, recent works have explored data-driven surrogates for IR drop estimation, leveraging machine learning and deep learning to predict full-chip maps orders of magnitude faster than traditional simulation~\cite{pao2020xgbir,chhabria2021iredge,xie2020powernet, fang2018machine}. Most existing methods, however, utilize power distribution or designer-extracted features as input and often neglect detailed PDN layout or inter-layer properties, limiting their ability to fully capture the complex spatial dependencies present in multi-layer power grids. Our main contributions are as follows:

\begin{itemize}
    \item We design a novel encoder–decoder architecture by integrating ConvNeXtV2 blocks~\cite{woo2023convnext} with attention gates from Attention U-Net~\cite{oktay2018attention}, enabling efficient multi-scale feature extraction and spatially adaptive skip connections.
    
    \item We propose a recursive two-stage channel attention mechanism, termed \textit{Weakness-Aware Channel Attention (WACA)}, which extends standard modules such as SE and CBAM by selectively enhancing weaker, complementary channels through sequential recalibration. This directly addresses channel imbalance caused by heterogeneous resistivity across PDN layers.
    
    \item WACA introduces no additional learnable parameters and can be easily integrated into existing attention modules, offering a lightweight and plug-and-play solution.
    
    \item Integrated into a ConvNeXtV2-based attention U-Net, our framework achieves state-of-the-art results on the ICCAD-2023 static IR drop benchmark reducing the mean absolute error by 61.1\% and increasing F1-score by 71.0\% compared to the contest winner~\cite{kadagala2023iccad}.
\end{itemize}


\section{Related Works}

\subsection{Feature-Driven Machine Learning}
Early approaches for IR drop estimation relied on feature-driven regressors, using designer-extracted metrics such as local power density, current flow, and wire resistance as inputs to classical machine learning models~\cite{pao2020xgbir,fang2018machine}. While these methods achieve fast inference, their lack of spatial awareness and generalization across layouts limits their practical adoption compared to recent deep learning techniques.

\subsection{Image-Based Deep Learning}
Convolutional neural networks (CNNs) have been widely applied to model IR drop as an image-to-image regression problem. PowerNet~\cite{xie2020powernet} uses a CNN-based local prediction framework leveraging cell-level features such as power components, signal timing, toggle rates, and spatial coordinates. IREDGe~\cite{chhabria2021iredge} extends this approach by adopting a U-Net architecture with physically relevant features including current maps, PDN density, and pad distance. However, these methods do not explicitly consider the 3D PDN layout, particularly the distinct roles and resistive properties of metal and via layers, limiting their modeling of inter-layer dependencies.

\subsection{Attention and Netlist-Aware Networks}
Recent advances integrate attention mechanisms and explicit netlist information to enhance spatial modeling and topology awareness. UnetPro~\cite{qi2024unetpro} and GLA-Inception U-Net~\cite{chen2024global} incorporate attention modules to focus on critical regions, while PDN-Net~\cite{zhao2024pdnnet} explicitly integrates netlist topology through graph neural networks. However, these approaches often handle multi-layer PDN inputs through simple concatenation, which may not adequately address the heterogeneous characteristics across different metal and via layers.

\subsection{Advanced Architectures with Comprehensive Features}
Several state-of-the-art architectures further improve IR drop prediction by incorporating richer input features and enhanced backbones. ICCAD-2023 contest winner~\cite{kadagala2023iccad} and CFIRSTNET~\cite{liu2024cfirstnet} both adopt ConvNeXtV2~\cite{woo2023convnext} based architectures, demonstrating the effectiveness of modern CNN backbones. CFIRSTNET defines a comprehensive multi-channel feature set by integrating spatial layout maps (pad distance and PDN density) and netlist-derived physical quantities (i.e. specifically resistance and current information). These are encoded as separate feature maps for each layer. Hypothetical IR drop maps generated from simplified analysis are also included as input channels. Similarly, CG-iAFFUNet~\cite{liu2024simultaneous} augments U-Net~\cite{ronneberger2015u} with iterative attention fusion modules, incorporating both netlist-based feature maps and approximate IR drop maps derived from fast numerical solvers.

\section{Method}

\begin{figure*}[ht!]
    \centering
    \includegraphics[width=0.95\textwidth]{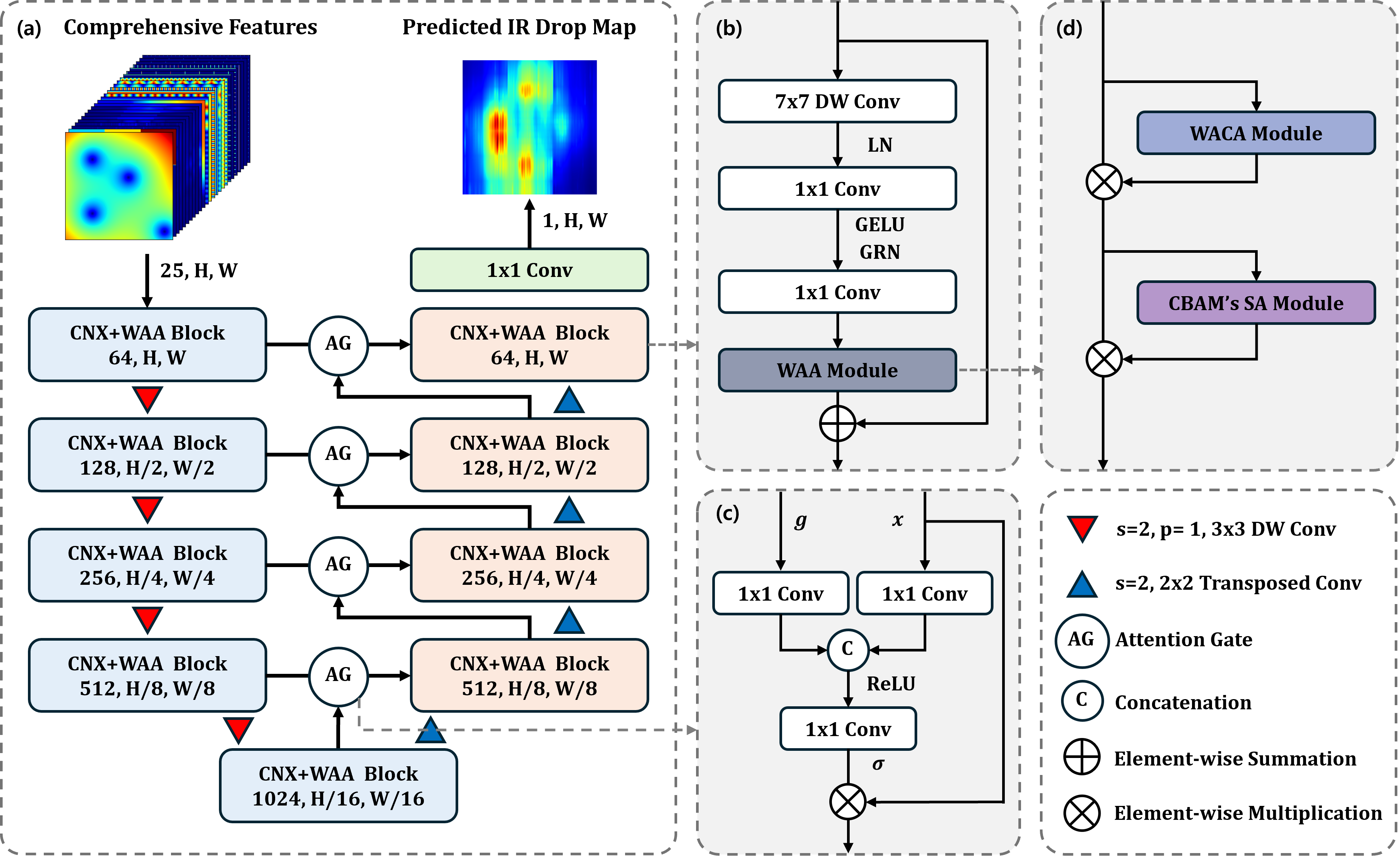}
    \caption{
        \textbf{Overall architecture of the proposed WACA-UNet.}
        (a) The main encoder–decoder network with attention gates and skip connections.
        (b) The CNX+WAA block, incorporating ConvNeXtV2 and the Weakness-Aware Attention (WAA) module.
        (c) The structure of the attention gate.
        (d) The overall structure of the WAA module, which includes the proposed Weakness-Aware Channel Attention (WACA) and CBAM’s spatial attention module.
        All diagram symbols are defined in the legend at the bottom right.
    }
    \label{fig:architecture}
\end{figure*}

Recent advances in IR drop prediction emphasize the importance of using advanced backbone architectures and comprehensive, multi-channel input features. CFIRSTNET~\cite{liu2024cfirstnet} demonstrated that integrating layout information and hypothetical IR drop maps significantly improves accuracy. However, existing models often overlook channel imbalance in multi-layer PDN inputs.

To address this limitation, we propose WACA, a channel attention mechanism that adaptively balances dominant and weak channels in feature representations. Our WACA-UNet, built on the ConvNeXtV2~\cite{woo2023convnext} based encoder–decoder architecture, leverages comprehensive feature sets for robust and precise IR drop prediction.

\subsection{Overall Architecture of WACA-UNet}
Fig.~\ref{fig:architecture} illustrates the overall architecture of the proposed WACA-UNet. The network adopts an encoder–decoder structure with skip connections and attention gates, enabling hierarchical feature extraction from comprehensive multi-channel input tensors.

We employ ConvNeXtV2~\cite{woo2023convnext} blocks as the backbone of our encoder–decoder architecture. ConvNeXtV2 is a state-of-the-art convolutional network that achieves competitive or superior performance to vision transformers~\cite{liu2021swin,dosovitskiy2020image} in large-scale image recognition, while maintaining the efficiency and inductive biases of traditional CNNs. Notably, ConvNeXtV2 can effectively capture global context without explicit attention mechanisms, owing to its advanced convolutional designs and Global Response Normalization (GRN) mechanism. Furthermore, it demonstrates strong performance even without pre-trained weights, enabling robust training from scratch on domain-specific tasks.

Each CNX+WAA block, as shown in Fig.~\ref{fig:architecture}(b), augments the ConvNeXtV2 backbone with a Weakness-Aware Attention (WAA) module to better aggregate and highlight salient features at every scale. Note that WAA represents the complete attention module combining multiple mechanisms, while WACA specifically refers to our novel channel attention component within WAA. We incorporate attention gates~\cite{oktay2018attention} into our skip connections (Fig.~\ref{fig:architecture}(c)) to selectively emphasize informative features from the encoder path. The attention gate mechanism computes attention coefficients that highlight salient features while suppressing less relevant activations, thereby improving the quality of feature fusion between encoder and decoder paths.

\begin{figure*}[ht]
    \centering
    \includegraphics[width=0.90\textwidth]{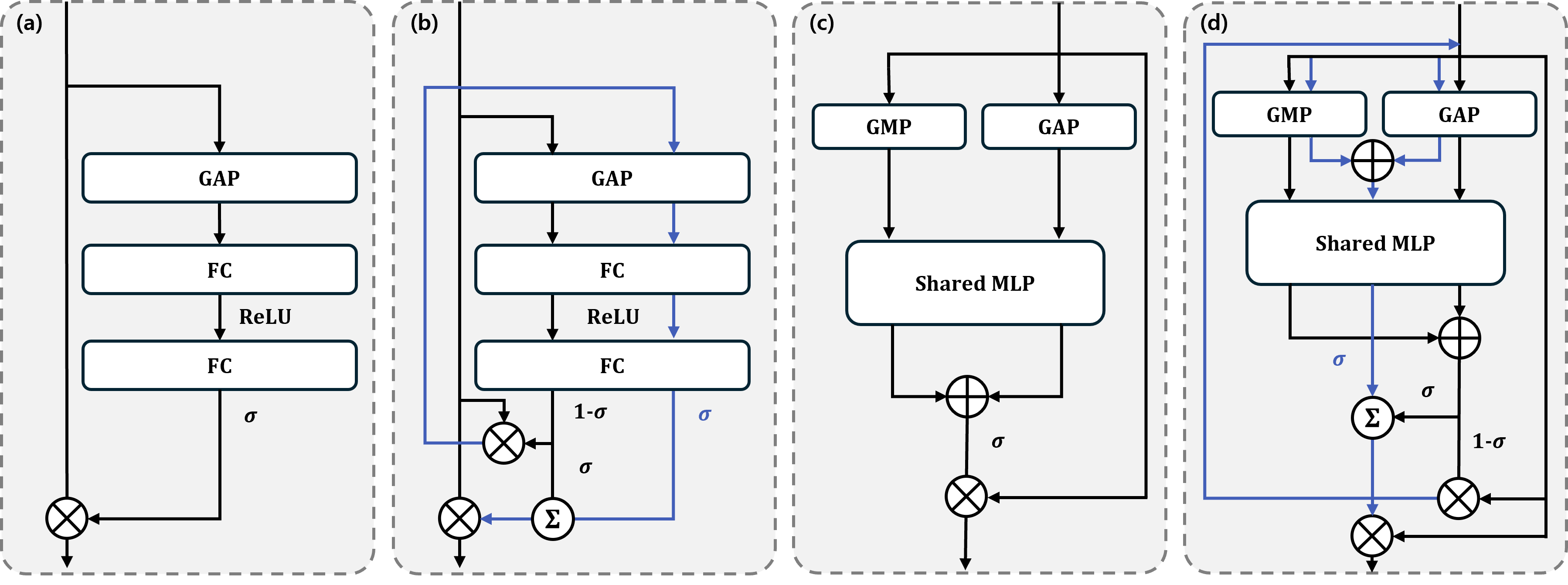}
    \caption{
        Comparison of channel attention modules.
        (a) The Channel Attention module of SENet.
        (b) The proposed WACA extension applied to SENet.
        (c) The Channel Attention module of CBAM.
        (d) The proposed Weakness-Aware Channel Attention (WACA) extension applied to CBAM. Blue lines denote additional gating paths introduced by WACA.
        \textbf{\textcircled{\scriptsize $\Sigma$}} 
        : element-wise weighted sum (equal weights $0.5$) 
        }
    \label{fig:waca}
\end{figure*}

\subsection{Weakness-Aware Channel Attention}

The design of our Weakness-Aware Channel Attention (WACA) mechanism is motivated by two key principles:  
(i) the \textbf{complementary information hypothesis}, which posits that weak channels may provide critical information absent from dominant ones, and  
(ii) \textbf{adaptive recalibration}, where a recursive two-stage gating mechanism enables dynamic balancing between strong and weak channel responses.

Fig.~\ref{fig:waca} compares baseline channel attention modules and our proposed WACA extensions. In (a), the standard Squeeze-and-Excitation Networks (SENet)~\cite{hu2018squeeze} channel attention is shown, which recalibrates channel-wise feature responses via global average pooling and fully connected layers. (b) illustrates our WACA-enhanced version of SENet, which introduces a second gating path and fusion operation (highlighted in blue) to emphasize weak channels. (c) depicts the Convolutional Block Attention Module (CBAM)~\cite{woo2018cbam}, where channel and spatial attention are applied sequentially. (d) presents WACA-CBAM, where we extend the channel attention stage with recursive gating. The additional WACA-specific logic is shown in blue.

Existing channel attention mechanisms, such as SENet and CBAM, primarily focus on identifying and amplifying dominant channels. However, in multi-layer PDN representations, this approach can exacerbate channel imbalance by further suppressing already weak but potentially informative channels. To address this limitation, we propose a WACA mechanism that explicitly models and enhances weak channels.

Rather than directly computing attention weights for channel enhancement, WACA employs a \textbf{recursive two-stage gating strategy} that operates via sequential channel suppression and recalibration using shared parameters. This process also acts as an implicit regularizer, discouraging over-reliance on dominant channels and promoting more robust, balanced feature learning.

Given input feature maps $\mathbf{X} \in \mathbb{R}^{C \times H \times W}$, WACA computes attention weights through a two-stage sequential process: first, suppression of dominant channels, and second, recalibration to enhance weak channels. We present two WACA variants based on different backbone attention mechanisms below:

\textbf{WACA-SE (SENet-based):}

    \textbf{Stage 1: Strong Channel Identification.}
    
    First, the standard SENet channel attention is computed:
    \begin{align}
        \mathbf{s}_1 = \mathrm{GAP}(\mathbf{X}), \quad
        \mathbf{a}_1 = \sigma(\mathrm{FC}_2(\delta(\mathrm{FC}_1(\mathbf{s}_1)))),
    \end{align}
    where $\mathrm{GAP}$ denotes global average pooling, $\delta(\cdot)$ is the ReLU activation, and $\sigma(\cdot)$ is the sigmoid function. $\mathbf{a}_1$ serves as the attention weights that primarily highlight dominant channels, as in the original SENet.

    \textbf{Stage 2: Weak Channel Enhancement.}
    To expose and recalibrate weak channels, we first suppress dominant channels by constructing a complementary weight $\mathbf{w}_1 = 1 - \mathbf{a}_1$ and use it to modulate the input features:
    \begin{align}
        \mathbf{s}_2 = \mathrm{GAP}(\mathbf{X} \odot \mathbf{w}_1), \quad
        \mathbf{a}_2 = \sigma(\mathrm{FC}_2(\delta(\mathrm{FC}_1(\mathbf{s}_2)))).
    \end{align}
    The second attention gate $\mathbf{a}_2$ is thus computed from features where dominant channels have been attenuated, allowing the model to focus on informative weak channels.
    
    \noindent
    \textbf{Stage 3: Adaptive Fusion.}
    The final attention output fuses both attention gates:
    \begin{align}
        \mathbf{y} = \mathbf{X} \odot (\alpha \mathbf{a}_1 + (1-\alpha)\mathbf{a}_2),
    \end{align}
    where $\alpha$ is a balancing parameter (set to $0.5$ in our experiments), and $\odot$ denotes channel-wise multiplication.

\textbf{WACA-CBAM (CBAM-based):}

\textbf{Stage 1: Multi-pooling Strong Channel Identification.}
Following CBAM, we apply both global average and max pooling, and combine their outputs through a shared MLP to obtain the first attention gate:
\begin{align}
    \mathbf{s}_{\mathrm{avg}} = \mathrm{GAP}(\mathbf{X}), \quad
    \mathbf{s}_{\mathrm{max}} = \mathrm{GMP}(\mathbf{X}), \\
    \mathbf{a}_1 = \sigma(\mathrm{MLP}(\mathbf{s}_{\mathrm{avg}}) + \mathrm{MLP}(\mathbf{s}_{\mathrm{max}})),
\end{align}
where $\mathrm{GMP}$ denotes global max pooling, and $\mathrm{MLP}(\cdot)$ is a two-layer shared MLP with ReLU activation.

\textbf{Stage 2: Weak Channel Enhancement.}
Analogous to WACA-SE, we suppress dominant channels by using the complementary weight $\mathbf{w}_1 = 1 - \mathbf{a}_1$ and apply both global pooling operations to the recalibrated features:
\begin{align}
    \mathbf{s}_w = \mathrm{GAP}(\mathbf{X} \odot \mathbf{w}_1) + \mathrm{GMP}(\mathbf{X} \odot \mathbf{w}_1), \\
    \mathbf{a}_2 = \sigma(\mathrm{MLP}(\mathbf{s}_w)),
\end{align}
where $\mathbf{a}_2$ now reflects attention over the previously suppressed, weak channels.

\textbf{Stage 3: Adaptive Fusion.}
The final attention output is obtained by adaptively combining both attention gates:
\begin{align}
    \mathbf{y} = \mathbf{X} \odot (\alpha \mathbf{a}_1 + (1-\alpha)\mathbf{a}_2).
\end{align}
As before, $\alpha$ is a balancing parameter set to 0.5.

WACA is integrated into our WAA module, which combines channel-wise recalibration with spatial attention. The WAA module follows the CBAM design principle but replaces the standard channel attention with our WACA mechanism:
\begin{equation}
    \mathrm{WAA}(\mathbf{X}) = \mathrm{SpatialAttention}(\mathrm{WACA}(\mathbf{X})).
\end{equation}
This design ensures that both channel imbalance and spatial feature distribution are simultaneously addressed, leading to more balanced and informative feature representations.

\section{Experiments}
\subsection{Experimental Setup}
\textbf{Dataset.}
Our experiments are conducted on the ICCAD-2023 contest benchmark~\cite{kadagala2023iccad}, which provides a comprehensive evaluation framework for IR drop prediction. The dataset consists of both synthetic and real circuit data synthesized using the open-source NanGate 45nm technology node~\cite{ajayi2019openroad}. Due to the limited availability of real training data, our training set comprises 2,100 synthetic cases: 2,000 cases from BeGAN's nangate45\_set1 and nangate45\_set2~\cite{chhabria2021began}, and 100 synthetic contest cases from the ICCAD-2023 dataset. The validation and test sets each consist of 10 real cases from the ICCAD-2023 contest, with the test set serving as the hidden-real evaluation. Following the approach of CFIRSTNET~\cite{liu2024cfirstnet}, we use the comprehensive 25-channel feature set as input, and the IR drop map as the ground truth target.

\textbf{Data Preprocessing and Normalization.}
All input data (training, validation, and test) are resized to a resolution of $384\times 384$ pixels for model processing, with Lanczos interpolation used for all resizing operations to preserve image quality. During training, both input features and target IR drop maps are processed at $384\times 384$ resolution, and loss is computed at this resolution. During validation and testing, input features are resized to $384\times 384$ for model inference, and the resulting $384\times 384$ predictions are then resized back to the original resolution for evaluation against the original ground truth. For input features, we apply z-score normalization using the mean and standard deviation computed from the training set, and apply the same statistics to both validation and test sets. The target IR drop values are converted from volts to millivolts (by multiplying by 1,000) without further normalization.

To enhance model generalization and robustness, we apply several data augmentation techniques during training, including horizontal and vertical flips as well as $90^\circ$, $180^\circ$, and $270^\circ$ rotations. These augmentations are particularly effective for circuit layout data, as they preserve electrical properties while increasing data diversity.

\textbf{Training Configuration.}
The proposed model is implemented in PyTorch, and all experiments are conducted on a single NVIDIA RTX A6000 GPU. We train the model for 500 epochs using a batch size of 4, with the AdamW~\cite{loshchilov2017decoupled} optimizer (learning rate of $4\times 10^{-5}$ and weight decay of $1\times 10^{-3}$). The learning rate is scheduled with cosine annealing~\cite{loshchilov2016sgdr} for stable convergence. The loss function is a composite of the Structural Similarity Index Measure (SSIM) loss~\cite{wang2004image}, Huber loss~\cite{huber1992robust}, and focal frequency loss~\cite{jiang2021focal}, designed to capture both spatial structure and pixel-level accuracy. During training, the model checkpoint with the best validation F1-score is selected for final evaluation.

\textbf{Evaluation Metrics.}
Following the ICCAD-2023 contest~\cite{kadagala2023iccad} evaluation protocol, we evaluate model performance using two primary metrics: mean absolute error (MAE) and F1-score for hotspot detection. The MAE measures the average absolute difference between predicted and ground truth IR drop values across all pixels:
\begin{equation}
    \mathrm{MAE} = \frac{1}{N} \sum_{i=1}^N |y_i - \hat{y}_i|,
\end{equation}
where $y_i$ and $\hat{y}_i$ denote the ground truth and predicted IR drop values for the $i$-th pixel, and $N$ is the total number of pixels.

For hotspot detection, we define hotspots as regions where IR drop values exceed $90\%$ of the maximum value in each map, i.e., $\mathrm{Hotspot} = \{ i \mid y_i \geq 0.9 \cdot y_{\max} \}$, by following the same method in \cite{kadagala2023iccad}. The F1-score is computed as
\begin{equation}
    \mathrm{F1} = \frac{2\,\mathrm{TP}}{2\,\mathrm{TP} + \mathrm{FP} + \mathrm{FN}},
\end{equation}
where $\mathrm{TP}$, $\mathrm{FP}$, and $\mathrm{FN}$ are the numbers of true positives, false positives, and false negatives, respectively, for hotspot prediction. The F1-score provides a balanced evaluation of both precision and recall in hotspot identification, which is critical for highlighting regions that require design optimization.

\subsection{Experimental Results}
\begin{figure}[ht]
    \centering
    \includegraphics[width=0.9\linewidth]{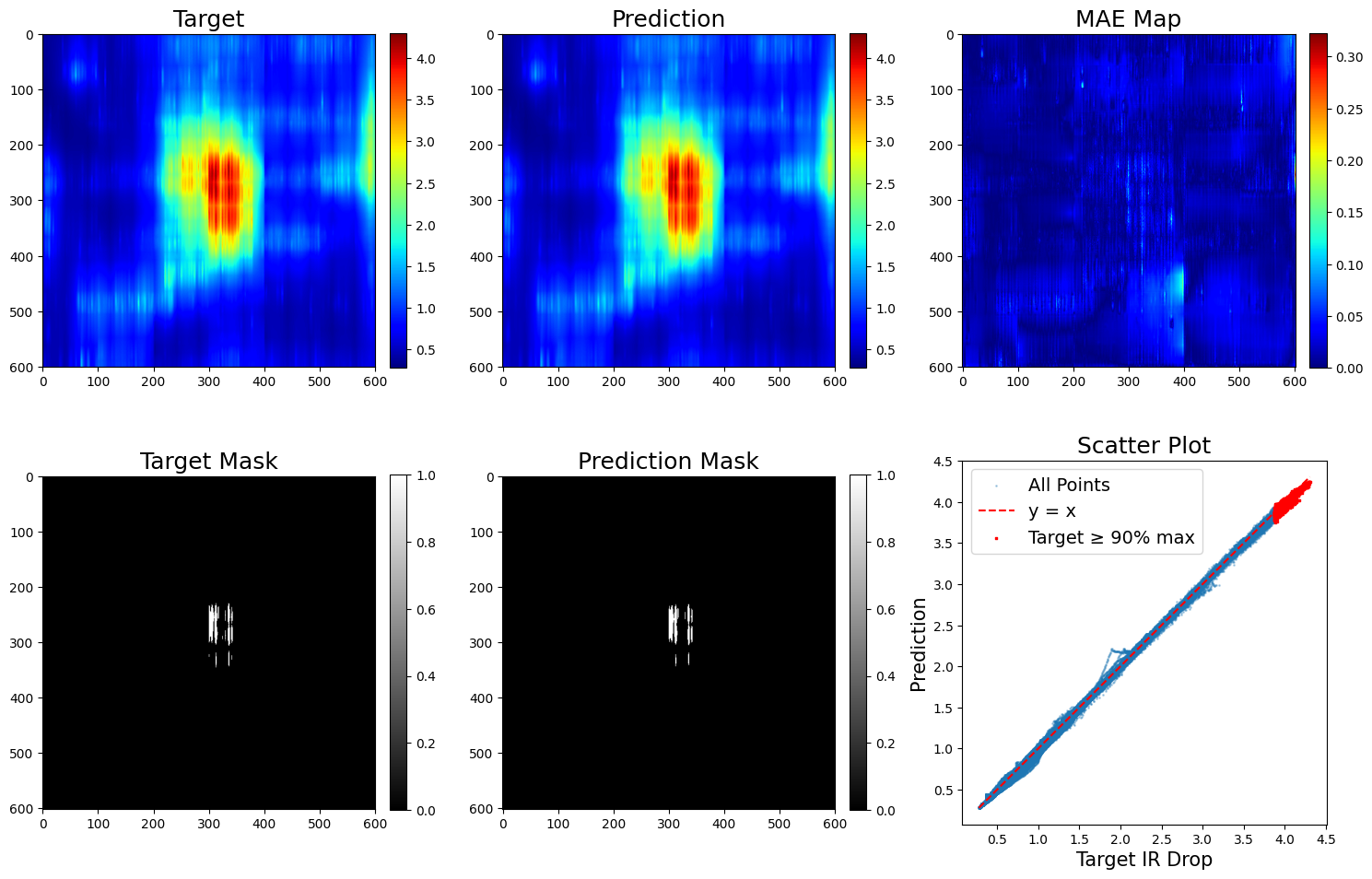}
        \caption{
    Qualitative results for IR drop prediction on \textbf{testcase7}. Top: target, prediction, and MAE map. Bottom: target mask, prediction mask, and scatter plot of predicted vs. target IR drop. Red dots indicate hotspot pixels.
    }
    \label{fig:inference}
\end{figure}

This section provides both quantitative and qualitative evaluation results of the proposed WACA-UNet on the ICCAD-2023~\cite{kadagala2023iccad} benchmark.
Fig.~\ref{fig:inference} provides a qualitative example of IR drop prediction using WACA-UNet.
The predicted IR drop map closely matches the ground-truth distribution, and WACA-UNet accurately identifies hotspot regions with high spatial accuracy.

To support these qualitative observations, Table~\ref{tab:results} quantitatively compares the performance of WACA-UNet against recent state-of-the-art methods, including IREDGe~\cite{chhabria2021iredge}, the ICCAD-2023 contest winner~\cite{kadagala2023iccad}, CFIRSTNET~\cite{liu2024cfirstnet}, and CG-iAFFUNet~\cite{liu2024simultaneous}. The table also includes the runtime (RT) in seconds for each model, measured on CPU, to assess inference efficiency.
While prior approaches such as CFIRSTNET and CG-iAFFUNet demonstrate strong performance on selected testcases (e.g., testcase13, testcase14), their effectiveness varies significantly across different design scenarios.
In contrast, WACA-UNet consistently delivers competitive results across the entire benchmark, achieving the lowest average MAE of 0.0524 and the highest average F1 score of 0.778 on the ICCAD-2023 dataset.

Despite achieving the highest average F1 score, WACA-UNet maintains a low standard deviation of 0.140 in F1 across all testcases, indicating consistent prediction quality.
This standard deviation is comparable to CFIRSTNET (0.139) and lower than CG-iAFFUNet (0.159), underscoring the robustness of WACA-UNet in diverse design scenarios.
In terms of MAE, WACA-UNet achieves the lowest error in 4 out of 10 testcases. This corresponds to a 1.7\% reduction compared to CFIRSTNET, a 24.2\% reduction compared to CG-iAFFUNet, and a 61.1\% reduction compared to the ICCAD-2023 contest winner.
For F1 score, WACA-UNet also achieves the highest value in 4 out of 10 testcases. This represents a 7.6\% improvement over CFIRSTNET, a 5.6\% improvement over CG-iAFFUNet, and a substantial 71.0\% improvement over the contest winner.
In addition to predictive accuracy, WACA-UNet achieves a reasonable average runtime of 1.39 seconds per testcase. Although CFIRSTNET achieves the shortest runtime among compared methods, WACA-UNet maintains high accuracy with practical inference speed.
These results highlight the robustness and generalization capability of the proposed weakness-aware attention mechanism, which enables stable and reliable IR drop prediction under diverse and complex circuit conditions.

\begin{table}[ht]
\centering
\renewcommand{\arraystretch}{1.2}
\resizebox{\textwidth}{!}{%
\begin{threeparttable}
\begin{tabular}{l
                r@{\hspace{4pt}}r@{\hspace{4pt}}r
                r@{\hspace{4pt}}r@{\hspace{4pt}}r
                r@{\hspace{4pt}}r@{\hspace{4pt}}r
                r@{\hspace{4pt}}r@{\hspace{4pt}}r
                r@{\hspace{4pt}}r@{\hspace{4pt}}r}
\toprule
\multirow{2}{*}{Testcase} 
  & \multicolumn{3}{c}{\textbf{IREDGe~\cite{chhabria2021iredge}}} 
  & \multicolumn{3}{c}{\textbf{Contest Winner~\cite{kadagala2023iccad}}}
  & \multicolumn{3}{c}{\textbf{CFIRSTNET~\cite{liu2024cfirstnet}}}
  & \multicolumn{3}{c}{\textbf{CG-iAFFUNet~\cite{liu2024simultaneous}}}
  & \multicolumn{3}{c}{\textbf{WACA-UNet (Ours)}} \\
\cmidrule(lr){2-4}\cmidrule(lr){5-7}\cmidrule(lr){8-10}\cmidrule(lr){11-13}\cmidrule(lr){14-16}
  & MAE $\downarrow$ & F1  $\uparrow$  & RT $\downarrow$ 
  & MAE $\downarrow$ & F1  $\uparrow$  & RT $\downarrow$
  & MAE $\downarrow$ & F1  $\uparrow$  & RT $\downarrow$
  & MAE $\downarrow$ & F1  $\uparrow$  & RT $\downarrow$
  & MAE $\downarrow$ & F1  $\uparrow$  & RT $\downarrow$ \\
\midrule
testcase7   & 0.6218 & 0.142 & 0.150 & 0.0656 & 0.783 & 7.996  & \underline{0.0177} & \underline{0.923} & 0.366 & 0.0419 & 0.704 & 7.22 & \textbf{0.0147} & \textbf{0.953} & 1.384 \\
testcase8   & 0.3845 & 0.419 & 0.149 & 0.0815 & 0.816 & 8.396  & \underline{0.0257} & \underline{0.916} & 0.367 & 0.0693 & 0.796 & 7.72 & \textbf{0.0215} & \textbf{0.964} & 1.447 \\
testcase9   & 0.4538 & 0     & 0.264 & 0.0406 & 0.589 & 11.417 & 0.0278 & 0.526 & 0.572 & \textbf{0.0225} & \textbf{0.768} & 11.95 & \underline{0.0241} & \underline{0.721} & 1.408 \\
testcase10  & 0.2426 & 0     & 0.262 & 0.0659 & \underline{0.532} & 11.270 & \underline{0.0459} & 0.464 & 0.551 & 0.0741 & 0.414 & 11.41 & \textbf{0.0402} & \textbf{0.575} & 1.327 \\
testcase13  & 0.2441 & 0     & 0.033 & 0.2068 & 0     & 5.452  & \textbf{0.0774} & \underline{0.680} & 0.204 & 0.1310 & \textbf{0.696} & 3.21 & \underline{0.0861} & 0.561 & 1.346 \\
testcase14  & 0.3138 & 0     & 0.033 & 0.4215 & 0     & 5.463  & 0.1895 & \underline{0.678} & 0.203 & \textbf{0.0945} & \textbf{0.757} & 3.36 & \underline{0.1518} & 0.667 & 1.488 \\
testcase15  & 0.1530 & 0     & 0.105 & 0.0968 & 0.088 & 8.137  & \textbf{0.0353} & \underline{0.733} & 0.327 & 0.0789 & 0.521 & 6.24 & \underline{0.0549} & \textbf{0.749} & 1.359 \\
testcase16  & 0.2675 & 0.258 & 0.102 & 0.1601 & 0.529 & 7.413  & \underline{0.0763} & 0.785 & 0.324 & 0.0896 & \textbf{0.902} & 5.76 & \textbf{0.0734} & \underline{0.865} & 1.308 \\
testcase19  & 1.0649 & 0 & 0.281 & 0.0905 & 0.501 & 11.905 & \textbf{0.0187} & 0.752 & 0.596 & 0.0423 & \textbf{0.845} & 13.75 & \underline{0.0366} & \underline{0.817} & 1.432 \\
testcase20  & 0.8204 & 0     & 0.279 & 0.1180 & 0.711 & 11.758 & \textbf{0.0191} & 0.773 & 0.576 & 0.0473 & \textbf{0.969} & 12.90 & \underline{0.0205} & \underline{0.910} & 1.405 \\
\midrule
\textbf{Average} 
            & 0.4566 & 0.082 & 0.166
            & 0.1347 & 0.455 & 8.921 
            & \underline{0.0533} & 0.723 & 0.409 
            & 0.0691 & \underline{0.737} & 8.35 
            & \textbf{0.0524} & \textbf{0.778} & 1.390 \\
\bottomrule
\end{tabular}
\begin{tablenotes}
\scriptsize
\item[*] IREDGe results are from the re-implementation reported in CFIRSTNET.
\end{tablenotes}
\end{threeparttable}
}
\vspace{0.5em}
\caption{
    Performance comparison on the ICCAD-2023 benchmark.
    Best results in bold, second-best underlined.
}
\label{tab:results}
\end{table}

\section{Analysis}

\subsection{Effects of WACA on Attention Layer}

\begin{figure}[ht]
    \centering
    \begin{minipage}{0.8\linewidth}
        \centering
        \includegraphics[width=\linewidth]{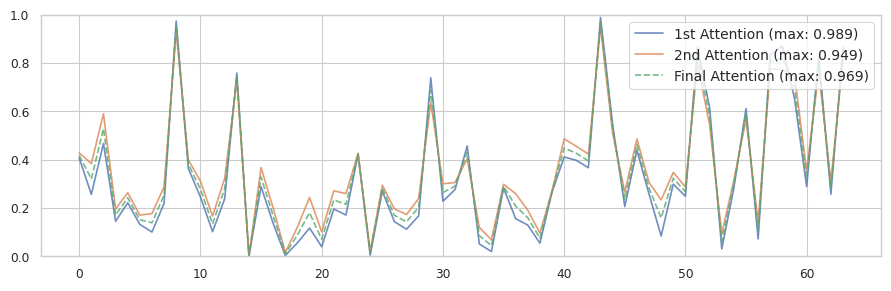}
        \includegraphics[width=\linewidth]{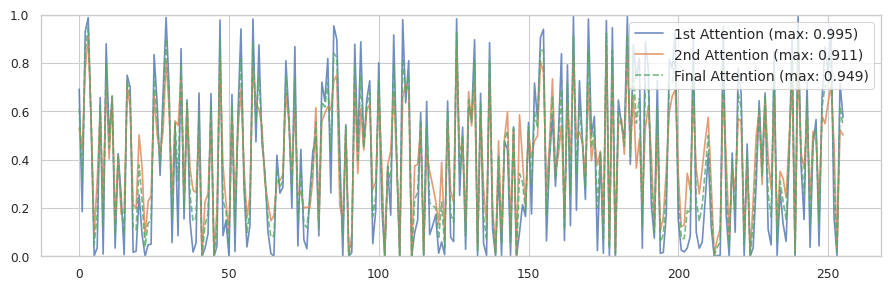}
    \end{minipage}
    \caption{
        Channel attention scores from the WACA module in the first encoder block (top, 64 channels) and third encoder block (bottom, 256 channels).
    }
    \label{fig:waca-attn-evolution}
\end{figure}

Fig.~\ref{fig:waca-attn-evolution} visualizes the channel attention scores at different stages within the WACA module for two representative encoder blocks. In both blocks, the first gating stage (blue) assigns high attention to dominant channels. After the second stage of recalibration (orange), the attention scores for weaker channels are adaptively increased, demonstrating the effectiveness of our recursive two-stage gating strategy. The final aggregated attention (green dashed) balances contributions from both stages, resulting in a more evenly distributed channel importance across the layers. This adaptive mechanism enables the model to avoid over-focusing on already strong channels while highlighting potentially informative weak channels, which is critical for robust IR drop prediction.

\subsection{Ablation Study}

Table~\ref{tab:ablation-attention} reports ablation results of two different channel attention modules integrated into the ConvNeXtV2~\cite{woo2023convnext} backbone and the impact of the proposed WACA mechanism. The baseline model without channel attention achieves an MAE of 0.0588 and F1-score of 0.670. Incorporating standard SE~\cite{hu2018squeeze} and CBAM~\cite{woo2018cbam} modules leads to moderate improvements; SE achieves MAE of 0.0563 and F1-score of 0.714, while CBAM shows 0.0671 MAE and 0.704 F1. Introducing WACA yields further gains: WACA-SE attains F1-score of 0.756 with 0.0571 MAE, and WACA-CBAM achieves the best performance with MAE of 0.0524 and F1-score of 0.778. These results demonstrate that the recursive two-stage gating of WACA significantly boosts hotspot detection accuracy and overall IR drop prediction, outperforming conventional attention mechanisms.

\begin{table}[ht]
    \centering
    \small
    \begin{tabular}{lccc}
        \toprule
        \textbf{Method}                & \textbf{MAE} $\downarrow$ & \textbf{F1} $\uparrow$ & \textbf{RT (s)} $\downarrow$ \\
        \midrule
        Baseline (only CNX)         & 0.0588     & 0.670    & 1.1737 \\
        + SE~\cite{hu2018squeeze}                        & 0.0563     & 0.714    & 1.2455 \\
        + CBAM~\cite{woo2018cbam}                      & 0.0665     & 0.731    & 1.3615 \\\midrule
        + WACA-SE                   & 0.0571     & 0.756    & 1.3445 \\
        + WACA-CBAM                 & \textbf{0.0524}     & \textbf{0.778}    & 1.3904 \\
        \bottomrule
    \end{tabular}
    \vspace{0.5em}
    \caption{Ablation results for channel attention modules.}
    \label{tab:ablation-attention}
\end{table}

\section{Conclusions}
In this paper, we proposed a novel Weakness-Aware Channel Attention (WACA) mechanism to address channel imbalance in multi-layer PDN-based IR drop prediction. By integrating WACA into a ConvNeXtV2-based attention U-Net architecture, our method adaptively enhances weak channels through a recursive two-stage gating strategy, resulting in more balanced feature utilization. Extensive experiments on the ICCAD-2023 benchmark demonstrate that our approach achieves state-of-the-art performance in both IR drop value estimation and hotspot detection, while maintaining practical inference speed. The proposed framework offers a promising direction for accurate and efficient power integrity analysis in advanced VLSI designs.

\bibliographystyle{unsrt}
\bibliography{references}
\end{document}